\documentclass{article}
\usepackage{spconf,amsmath,graphicx}
\usepackage{graphicx}
\usepackage{multirow}
\usepackage{tabularx}
\usepackage[dvipsnames]{xcolor}
\usepackage{booktabs}
\usepackage{colortbl}
\usepackage{amsmath}
\usepackage{amsfonts}
\usepackage{amssymb}
\usepackage{bbding}
\usepackage{enumitem}
\definecolor{cvprblue}{rgb}{0.21,0.49,0.74}
\definecolor{maroonx}{RGB}{195,18,48}
\usepackage[colorlinks=True,
            linkcolor=maroonx,
            anchorcolor=blue,
            citecolor=cvprblue,
            ]{hyperref}
\setlist[itemize]{leftmargin=*}
\setlist[enumerate]{noitemsep,leftmargin=*,topsep=0em}
\newcommand{\revise}[1]{{\color{black}{#1}}}
\usepackage{fancyhdr}
\fancypagestyle{firstpagefooter}{
    \fancyhf{} 
    \fancyfoot[C]{\small © 2025 IEEE. Personal use of this material is permitted.  Permission from IEEE must be obtained for all other uses, in any current or future media, including reprinting/republishing this material for advertising or promotional purposes, creating new collective works, for resale or redistribution to servers or lists, or reuse of any copyrighted component of this work in other works.}

    \fancyhead{} 
}

\title{Spectral-Aware Global Fusion for RGB-Thermal Semantic Segmentation}
\name{Ce Zhang \quad Zifu Wan \quad Simon Stepputtis \quad Katia Sycara\quad Yaqi Xie\vspace{-10pt}}
\address{Robotics Institute, Carnegie Mellon University\\
\texttt{\small\{cezhang, zifuw, sstepput, katia, yaqix\}@cs.cmu.edu}\vspace{-5pt}}

\begin{document}
\ninept
\maketitle
\begin{abstract}
Semantic segmentation relying solely on RGB data often struggles in challenging conditions such as low illumination and obscured views, limiting its reliability in critical applications like autonomous driving. To address this, integrating additional thermal radiation data with RGB images demonstrates enhanced performance and robustness. However, how to effectively reconcile the modality discrepancies and fuse the RGB and thermal features remains a well-known challenge. In this work, we address this challenge from a novel spectral perspective. We observe that the multi-modal features can be categorized into two spectral components: low-frequency features that provide broad scene context, including color variations and smooth areas, and high-frequency features that capture modality-specific details such as edges and textures. Inspired by this, we propose the Spectral-aware Global Fusion Network (SGFNet) to effectively enhance and fuse the multi-modal features by explicitly modeling the interactions between the high-frequency, modality-specific features. 
Our experimental results demonstrate that SGFNet outperforms the state-of-the-art methods on the MFNet and PST900 datasets.
\end{abstract}
\begin{keywords}
RGB-T Semantic Segmentation, Multi-Modal Fusion, Spectral-Aware Feature Fusion
\end{keywords}
\section{Introduction}
\label{sec:intro}
\looseness=-1
Semantic segmentation, which involves pixel-level scene understanding, is crucial for how autonomous agents perceive and interact with their environment. 
It empowers autonomous driving vehicles to distinguish between roads, pedestrians, and obstacles in real-time, ensuring safe navigation through complex urban environments~\cite{feng2020deep,sun2019rtfnet,wan2025sigma}.
Recently, the limited reliability of RGB images in challenging deployment scenarios---such as occlusions and varying light conditions, which lead to over- or under-exposure---has prompted the increasing use of thermal data as complementary information to enhance segmentation performance~\cite{ha2017mfnet,deng2021feanet}.
Unlike RGB data that uses visible light, thermal data captures thermal radiation, offering superior object detection capabilities in complex environments and uncovering elements that are invisible to RGB sensors.

\begin{figure}[t]
\centering
\includegraphics[width=\linewidth]{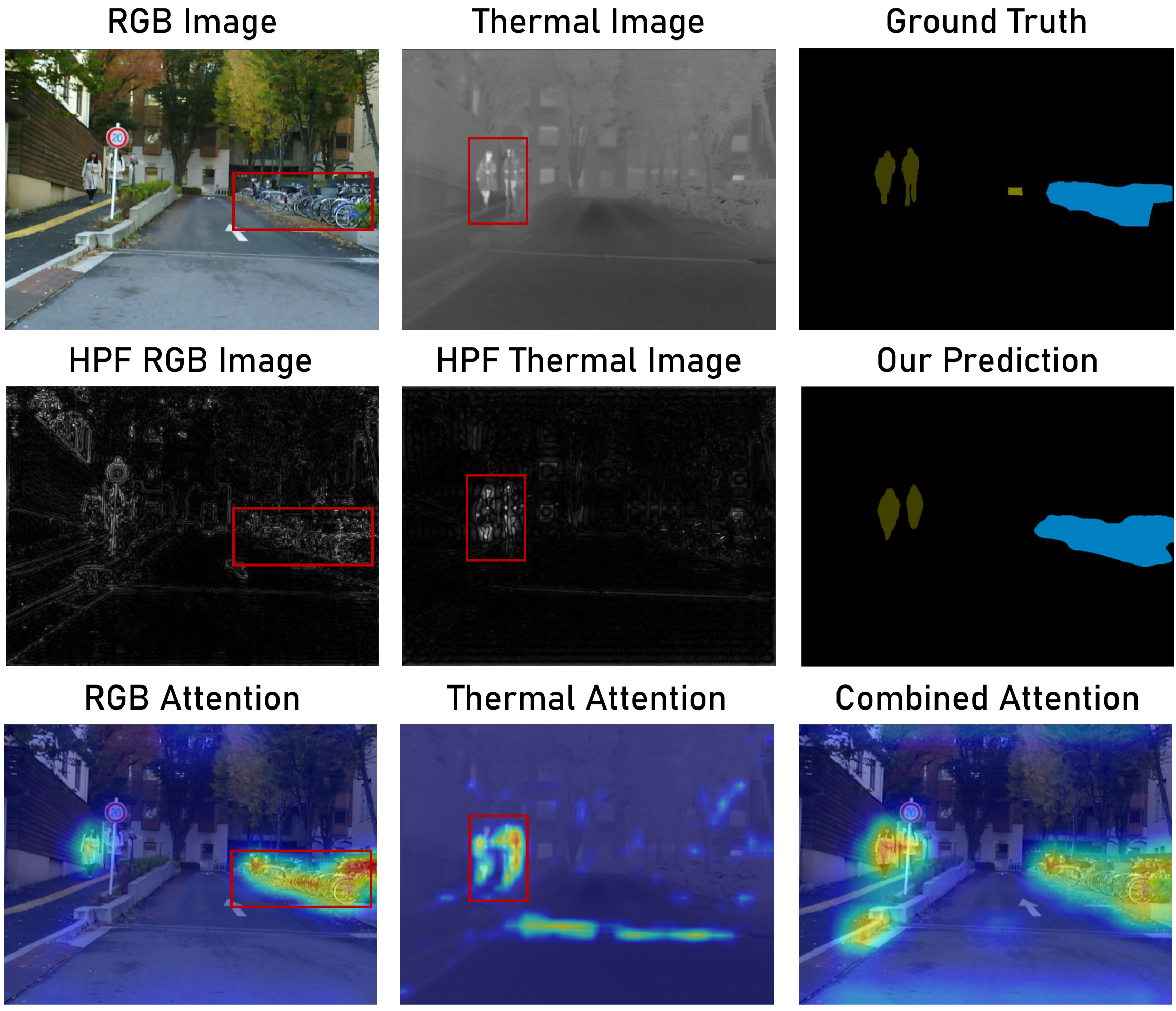}
\vspace{-18pt}
\caption{\textbf{Interpreting an image pair from a spectral perspective.} The high-frequency components are extracted by high-pass filtering (HPF) of the image's Fourier spectrum. The attention maps are visualized using Grad-CAM~\cite{selvaraju2017grad}.
We demonstrate that the HPF images capture texture and edge details that are specific to each modality. In our proposed SGFNet, we explicitly consider the interaction among these high-frequency components to effectively fuse multi-modal features and enhance segmentation performance.}
\label{fig:intro}
\vspace{-10pt}
\end{figure}

However, the significant variations between RGB and thermal modalities pose a notable challenge: effectively fusing the discriminative features from two modalities to enhance segmentation performance. Previous works have attempted to integrate thermal images by adding them as an additional dimension to RGB data, utilizing RGB-based architectures for segmentation~\cite{yu2018bisenet}, or by combining RGB and thermal features extracted from parallel encoders via simple addition~\cite{sun2019rtfnet,ha2017mfnet}.
More recently, approaches have shifted towards developing attention mechanisms to enhance the interaction of multi-modal features during fusion~\cite{deng2021feanet,liang2023explicit}.
Despite the promising results shown by these techniques, their effectiveness remains unresolved, as the complementary nature of RGB and thermal features is not fully understood or exploited.

\begin{figure*}[t]
\centering
\includegraphics[width=\linewidth]{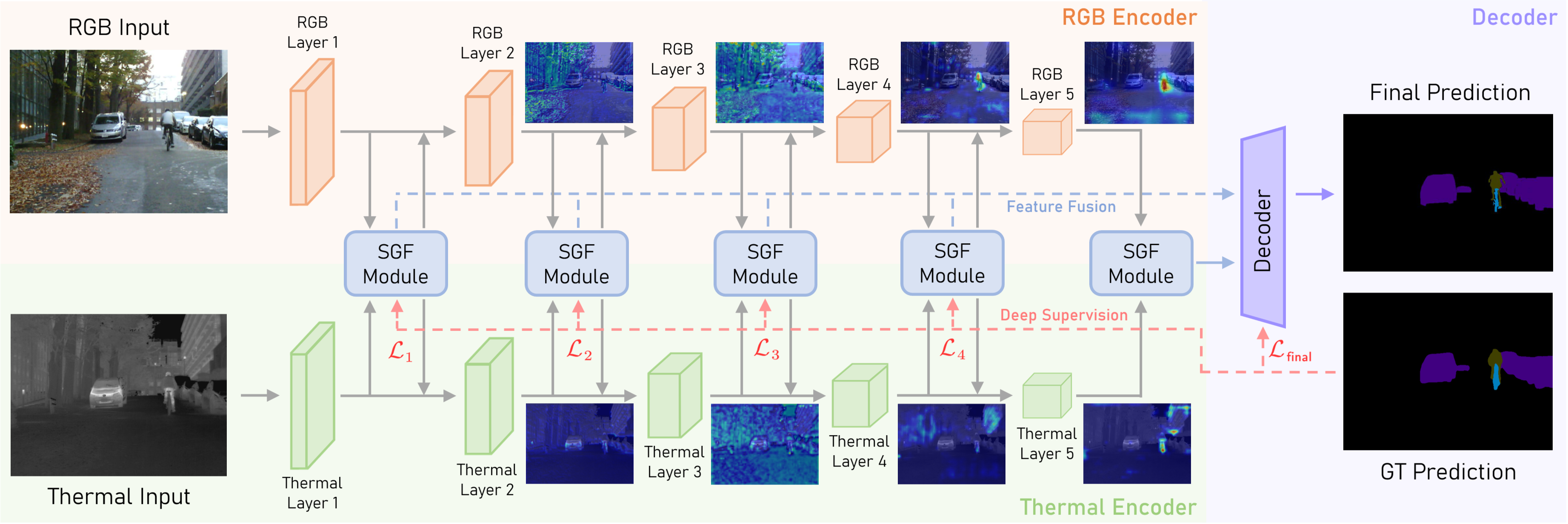} 
\vspace{-20pt}
\caption{\textbf{The overall framework of our SGFNet}. The SGF module is designed to effectively fuse multi-scale features derived from both RGB and thermal encoders. We also employ a deep supervision mechanism to supervise the preliminary prediction map at each scale.}
\label{fig:framework}
\vspace{-10pt}
\end{figure*}

\looseness=-1
To explore the complementary relationship between the two modalities more thoroughly, we ask a more in-depth research question: \textit{Which specific features within a pair of RGB and thermal data should be fused, and which should remain separate?}
In this work, we provide insights from a spectral perspective. Specifically, we observe two key points: (1) RGB and thermal features share similarities in their low-frequency responses, which contain lower-order statistics (\textit{e.g.}, brightness) and large-scale background features; (2) Conversely, their high-frequency information, which captures detailed textures and edges, remains unique to each modality.
For instance, consider the image pair in Figure \ref{fig:intro}. The high-frequency components of the RGB image successfully detect the bikes but fail to detect the pedestrians due to the occlusion caused by the traffic sign. In contrast, the high-frequency components of the thermal image accurately detect the pedestrians but struggle to detect the bikes due to their temperature being similar to the environment. 
This observation underscores the importance of prioritizing the fusion of high-frequency components to accurately segment distinct elements like pedestrians and bicycles in a scene.


\pagestyle{plain}

Motivated by the distinct properties of various spectral components, we introduce \textbf{S}pectral-aware \textbf{G}lobal \textbf{F}usion \textbf{Net}works (SGFNet), a novel approach designed to enhance and integrate multi-modal features more effectively. Unlike previous methods~\cite{deng2021feanet} that rely solely on maximum/average pooling for extracting representations from multi-modal features, SGFNet maps the features to the spectral domain and utilizes multi-spectral vectors that span the full frequency range within the spectral domain for richer representations. 
By explicitly enforcing the integration of high-frequency components between modalities, SGFNet demonstrates an improved performance in multi-modal fusion. To further enhance multi-modal interactions, we further introduce a global cross-attention module to capture long-range spatial dependencies in a cross-modal manner. 
Our extensive experiments on the MFNet~\cite{ha2017mfnet} and PST900~\cite{shivakumar2020pst900} datasets underscore SGFNet's superior capability in RGB-Thermal (RGB-T) semantic segmentation.

Our main contributions can be summarized as follows:
\begin{itemize}
    \item To the best of our knowledge, we propose the first spectral-aware method for RGB-T semantic segmentation, explicitly enforcing the integration of high-frequency components between modalities.
    \item We introduce a global cross-attention module to capture long-range spatial dependencies between modalities, improving the model's ability to integrate multi-modal data overarchingly.
    \item Our experiments on the MFNet and PST900 datasets demonstrate SGFNet's superior performance over state-of-the-art methods.
\end{itemize}

\section{Method}

\begin{figure*}[t]
\centering
\includegraphics[width=1\linewidth]{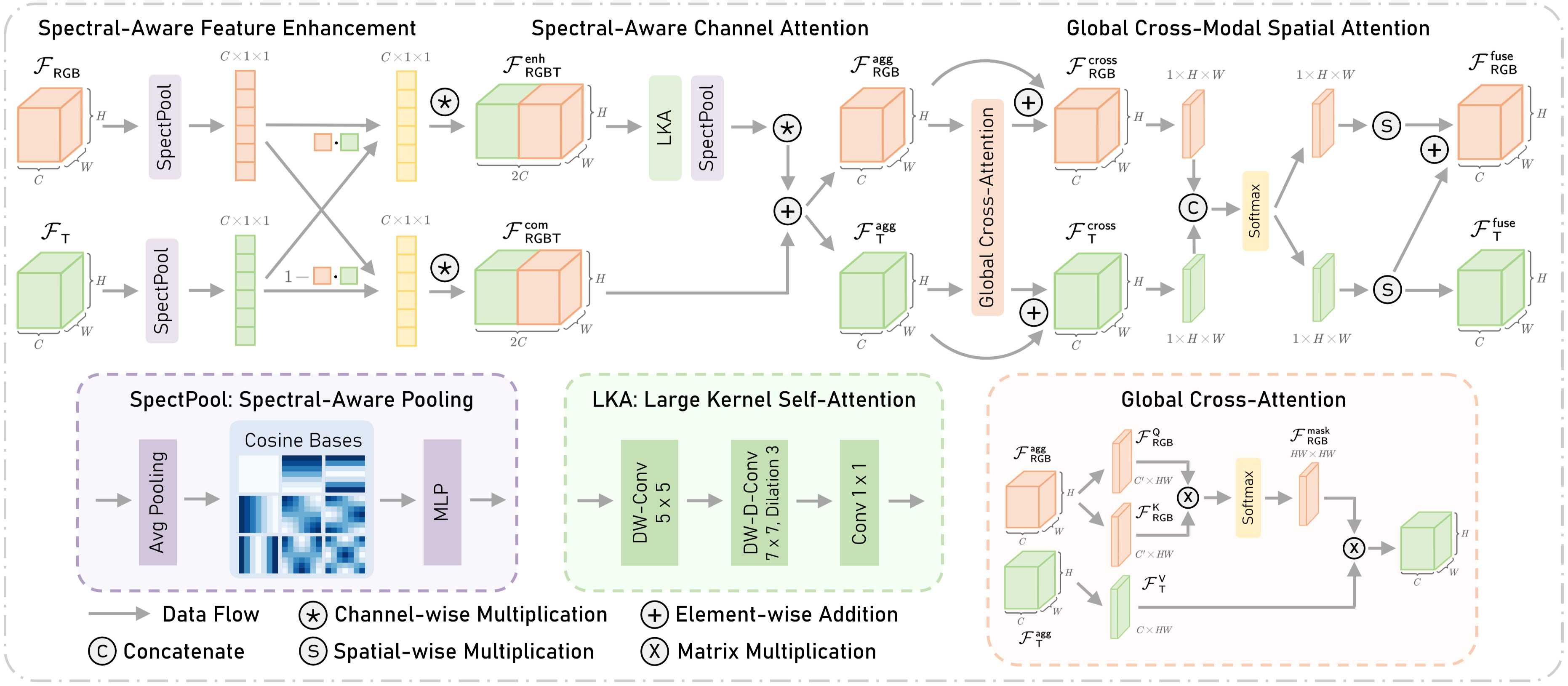}
\vspace{-20pt}
\caption{\revise{\textbf{Architecture of the proposed spectral-aware global fusion (SGF) module.} This module is composed of three sequential parts: spectral-aware feature enhancement, spectral-aware channel attention, and global cross-modal attention, respectively.}}
\vspace{-5pt}
\label{fig:SGF}
\vspace{-8pt}
\end{figure*}

\subsection{Spectral-Aware Feature Enhancement}
\label{subsec:enhance}
\noindent 
We utilize the ResNet-152~\cite{he2016deep} encoders to first extract features from both RGB and thermal images at a particular scale. These features, denoted as $\mathcal{F}_\mathsf{RGB/T}=f_{\mathsf{ResNet}}(\mathcal{I}_\mathsf{RGB/T})$, serve as the inputs for the SGF module.
As shown in Figure \ref{fig:SGF}, to enhance the interaction of the extracted features, we jointly assign weights to each channel of both modalities based on a spectral-aware channel-wise process. 

As we introduced in Section \ref{sec:intro}, the high-frequency features carry modality-specific finer details, which can enhance the segmentation accuracy of the model.
Therefore, to capture high-frequency components of the multi-modal features and enrich the representations of different channels, we introduce a set of cosine bases of two-dimensional Discrete Cosine Transforms (DCT)~\cite{qin2021fcanet}:
\begin{equation}
\setlength\abovedisplayskip{10pt}
\setlength\belowdisplayskip{10pt}
    \mathcal{B}_{f_h, f_w}^{hw} = \cos\left(\left(h+\frac{1}{2}\right)\pi \frac{f_h}{H'} \right) \cdot \cos\left(\left(w+\frac{1}{2}\right)\pi \frac{f_w}{W'} \right),
\end{equation}
in which $H'$ and $W'$ denote the size of the cosine basis, $f_h \in \{ 0, 1, \cdots, H'-1\}$, $f_w \in \{ 0, 1, \cdots, W'-1\}$ denotes the frequency along two axes, and $\mathcal{B}_{f_h, f_w}^{hw}$ represents the entry at position $(h, w)$ in the cosine basis $\mathcal{B}_{f_h, f_w}$.

To ensure compatibility between the input feature maps and the cosine bases, we begin by resizing the feature dimensions to $H' \times W'$ through average pooling. Then, we divide the input features $\mathcal{F}_\mathsf{RGB}$ and $\mathcal{F}_\mathsf{T}$ into $N$ groups along the channel dimension. These groups are denoted as $\mathcal{F}_\mathsf{RGB}^{i}$, $\mathcal{F}_\mathsf{T}^i \in \mathbb{R}^{(C/N) \times H' \times W'}$, where $i\in \{ 0, 1, \cdots, N-1\}$. For each group, we assign a unique cosine basis with frequency pair ${f_h^i, f_w^i}$. The discrete cosine transform (DCT) is then calculated by element-wise multiplication as follows:
\begin{equation}
\label{eq:dct}
\setlength\abovedisplayskip{3pt}
\setlength\belowdisplayskip{3pt}
\mathcal{S}^i_\mathsf{RGB/T} = \sum_{h=0}^{H'-1}\sum_{w=0}^{W'-1} \mathcal{F}_\mathsf{RGB/T}^{i, hw} \mathcal{B}_{f_h^i, f_w^i}^{hw},
\end{equation}
where $\mathcal{S}$ denotes a scalar representation of a particular channel corresponding to a specific DCT frequency component. Notably, when we select the lowest frequency pair, \textit{i.e.}, $f_h^i = 0$ and $f_w^i = 0$, the DCT simplifies to global average pooling (GAP) since $\mathcal{B}_{0, 0}^{hw} = 1$.

By concatenating these DCT components, we construct a multi-spectral vector that provides a comprehensive representation of each channel. Utilizing these vectors, we then derive a channel activation score by inputting them into a multi-layer perception (MLP):
\begin{align}
\label{eq:computeq}
\setlength\abovedisplayskip{3pt}
\setlength\belowdisplayskip{3pt}
\mathcal{Q}_\mathsf{RGB/T} &= f_{\mathsf{MLP}}\left([\mathcal{S}^0_\mathsf{RGB/T}, \mathcal{S}^1_\mathsf{RGB/T}, \cdots, \mathcal{S}^{N-1}_\mathsf{RGB/T}]\right)
\end{align}

We jointly enhance the features using the number of channels multiplied by the two activation scores $\mathcal{Q} = C \cdot \mathcal{Q}_\mathsf{RGB} \cdot \mathcal{Q}_\mathsf{T}$:
\begin{equation}
\label{eq:activation}
\setlength\abovedisplayskip{3pt}
\setlength\belowdisplayskip{3pt}
\mathcal{F}_\mathsf{RGB/T}^{\mathsf{enh}} = \mathcal{F}_\mathsf{RGB/T} \otimes\mathsf{Sigmoid}(\mathcal{Q}), 
\end{equation}
where $\otimes$ represents channel-wise multiplication. 
To prevent information loss, we incorporate a complementary branch to retain features that might be diminished by the enhancement, defined as:
\begin{equation}
\setlength\abovedisplayskip{3pt}
\setlength\belowdisplayskip{3pt}
\mathcal{F}_\mathsf{RGB/T}^{\mathsf{com}} = \mathcal{F}_\mathsf{RGB/T} \otimes \mathsf{Sigmoid}(1 - \mathcal{Q}).
\end{equation}

\subsection{Spectral-Aware Channel Attention}
\label{subsec:channel}
\noindent 
 In this section, we explicitly consider the interactions of the enhanced multi-modal features by assigning different weights across all $2C$ channels from both modalities.

We denote the concatenated enhanced feature maps of two modalities as $\mathcal{F}_\mathsf{RGBT}^{\mathsf{enh}} = [\mathcal{F}_\mathsf{RGB}^{\mathsf{enh}}, \mathcal{F}_\mathsf{T}^{\mathsf{enh}}]$, and similarly define  $\mathcal{F}_\mathsf{RGBT}^{\mathsf{com}} = [\mathcal{F}_\mathsf{RGB}^{\mathsf{com}}, \mathcal{F}_\mathsf{T}^{\mathsf{com}}]$.
For the enhanced feature maps $\mathcal{F}_\mathsf{RGBT}^{\mathsf{enh}}$, we introduce a spectral-aware interaction module to determine which channels across the two modalities are most beneficial for the final prediction. To capture long-range dependencies effectively, we initially employ a large kernel attention~\cite{guo2023visual} module to extract global self-attention. In this process, we implement a series of convolutions sequentially: first, a $5\times 5$ depth-wise convolution, followed by a $7\times 7$ depth-wise convolution with dilation 3, and finally, a $1\times 1$ convolution. The entire process can be formally written as
\begin{equation}
\setlength\abovedisplayskip{3pt}
\setlength\belowdisplayskip{3pt}
\mathcal{F}_\mathsf{RGBT}^{\mathsf{LKA}} = f_{\mathsf{conv}}^{1\times1}\!\left(f_{\mathsf{dwconv}}^{7\times7, \,\mathsf{d}3}\!\left(f_{\mathsf{dwconv}}^{5\times5}\!\left(\mathcal{F}_\mathsf{RGBT}^{\mathsf{enh}}\right)\right)\right).
\end{equation}

We similarly calculate the spectral-aware channel activation score $\mathcal{Q}_{\mathsf{RGBT}}^\mathsf{enh}$ of each channel for $\mathcal{F}_\mathsf{RGBT}^{\mathsf{LKA}}$ following Equations~(\ref{eq:dct}) and (\ref{eq:computeq}). The resulting multi-modal features from this interaction module are obtained by weighting the original $\mathcal{F}_\mathsf{RGBT}^{\mathsf{enh}}$ with the channel-wise activation score, which can be represented by
\begin{equation}
\setlength\abovedisplayskip{3pt}
\setlength\belowdisplayskip{3pt}
\mathcal{F}_\mathsf{RGBT}^{\mathsf{enh'}} = \mathcal{F}_\mathsf{RGBT}^{\mathsf{enh}} \otimes \mathsf{Sigmoid}(\mathcal{Q}_{\mathsf{RGBT}}^\mathsf{enh}).
\end{equation}

The cross-modal features in both the complementary and the enhancement branches are then aggregated:
\begin{equation}
\setlength\abovedisplayskip{3pt}
\setlength\belowdisplayskip{3pt}
\mathcal{F}_\mathsf{RGBT}^{\mathsf{agg}} = \mathcal{F}_\mathsf{RGBT}^{\mathsf{enh'}} + \mathcal{F}_\mathsf{RGBT}^{\mathsf{com}}.
\end{equation}

\vspace{-10pt}
\subsection{Global Cross-Modal Spatial Attention}
\label{subsec:spatial}
\looseness=-1 Having acquired the aggregated feature maps $\mathcal{F}_\mathsf{RGB/T}^{\mathsf{agg}}$ through channel attention, the next critical step is to complementarily merge the cross-modal features for pixels at varied locations.
Drawing inspiration from the efficacy of the self-attention mechanism~\cite{vaswani2017attention}, we have developed a global cross-attention module as shown in Figure~\ref{fig:SGF}. This module is designed to account for the interactions between every pair of pixels, thereby extending attention to encompass global spatial relationships to more effectively merge cross-modal features.

More specifically, we first apply $1\times 1$ convolutions to produce three matrices given the aggregated RGB/thermal feature map. These matrices function as the query ($\mathsf{Q}$), key ($\mathsf{K}$), and value ($\mathsf{V}$) components, respectively.
This process can be represented as
\begin{equation}
\setlength\abovedisplayskip{3pt}
\setlength\belowdisplayskip{3pt}
\mathcal{F}_\mathsf{RGB/T}^{\mathsf{Q/K/V}} = \mathsf{Reshape} \left(f_{\mathsf{conv}}^{1\times1}\left(\mathcal{F}_\mathsf{RGB/T}^{\mathsf{agg}}\right)\right).
\end{equation}
Specifically, to mitigate computational overhead, the $1\times 1$ convolutions utilized for generating the query and key matrices are designed with a reduced channel count, set to $C'=C/8$, whereas the convolutions for the value matrix retain the original $C$ channels. We reshape these matrices to achieve the desired dimensions, then compute the enhanced thermal feature map by cross-attention:
\begin{align}
\label{eq:cross}
\setlength\abovedisplayskip{0pt}
\setlength\belowdisplayskip{0pt}
\mathcal{F}_\mathsf{RGB}^\mathsf{mask} &= \mathsf{Softmax}\left({\mathcal{F}_\mathsf{RGB}^\mathsf{Q}}^{\top} \mathcal{F}_\mathsf{RGB}^\mathsf{K}\right), \\
\mathcal{F}_\mathsf{T}^\mathsf{cross} &= {\mathcal{F}_\mathsf{T}^\mathsf{V}}^{\top} \mathcal{F}_\mathsf{RGB}^\mathsf{mask} +  \mathcal{F}_\mathsf{T}^{\mathsf{agg}},
\end{align}
where the $\mathsf{Softmax}$ function is applied along the columns. In a similar manner, $\mathcal{F}_\mathsf{RGB}^\mathsf{cross}$ can be symmetrically derived using Equation~(\ref{eq:cross}).

Subsequently, we calculate the global cross-modal spatial attention to integrate the multi-modal features:
\begin{equation}
\setlength\abovedisplayskip{3pt}
\setlength\belowdisplayskip{3pt}
\mathcal{F}_\mathsf{RGB/T}^{\mathsf{att}} = \mathcal{F}_\mathsf{RGB/T}^{\mathsf{agg}} \odot\mathsf{Softmax}\left(f_{\mathsf{conv}}^{1\times1}\left(\mathcal{F}_\mathsf{RGB/T}^{\mathsf{cross}}\right)\right),
\end{equation}
where $\odot$ denotes element-wise spatial multiplication.

Finally, to obtain the ultimate fused feature map, we add the thermal feature map to the RGB stream:
\begin{equation}
\setlength\abovedisplayskip{3pt}
\setlength\belowdisplayskip{3pt}
\revise{
\mathcal{F}_\mathsf{RGB}^{\mathsf{fuse}} = \mathcal{F}_\mathsf{RGB}^{\mathsf{att}} + \mathcal{F}_\mathsf{T}^{\mathsf{att}}, \,
\mathcal{F}_\mathsf{T}^{\mathsf{fuse}} = \mathcal{F}_\mathsf{T}^{\mathsf{att}}.}
\end{equation}

\vspace{-10pt}
\subsection{Deep Supervision}
\label{subsec:deep}
\noindent After extracting and fusing the multi-modal features, we adopt the cascaded decoder proposed by BBS-Net~\cite{fan2020bbs} to further integrate these multi-level features and deliver final prediction $\mathcal{M}_{\mathsf{final}}$. The loss function for evaluating this final prediction is formulated as: 
\begin{equation}
\label{eq:loss1}
\setlength\abovedisplayskip{3pt}
\setlength\belowdisplayskip{3pt}
\mathcal{L}_{\mathsf{final}} = \mathsf{Loss}\left(\mathcal{M}_{\mathsf{final}}, \mathcal{M}_{\mathsf{GT}}\right),
\end{equation}
where $\mathcal{M}_{\mathsf{GT}}$ is the ground-truth segmentation map.

Additionally, within the SGF module, we also predict a preliminary prediction map based on the fused feature $\mathcal{F}_\mathsf{RGB}^{\mathsf{fuse}}$. 
To obtain more accurate segmentation results from multi-level features, we introduce an additional loss function to supervise the early-layer feature generation.
This function is applied to the preliminary results, specifically after the initial four layers: 
\begin{equation}
\label{eq:loss2}
\setlength\abovedisplayskip{3pt}
\setlength\belowdisplayskip{3pt}
\mathcal{L}_i = \mathsf{Loss}\left(\mathcal{M}_{\mathsf{pred}}^i, \mathcal{M}_{\mathsf{GT}}\right), i\in \{1, 2, 3, 4\}.
\end{equation}
The total loss function is then calculated as the sum of the individual loss components in Equations~(\ref{eq:loss1}) and (\ref{eq:loss2}).

\begin{table*}[t!]
  \centering
  \small
  \renewcommand{\arraystretch}{1.15}
  \renewcommand{\tabcolsep}{5pt}
  
  \caption{\looseness=-1
  \textbf{Quantitative performance comparisons with the state-of-the-art methods on the MFNet~\cite{ha2017mfnet} dataset}. The best results for each metric are in \textbf{bold}, while the second-best results are \underline{underlined}. \revise{$^\dagger$These methods utilize the more advanced SegFormer~\cite{xie2021segformer} backbone.}}
  \label{tab:mfnet}
\resizebox{1\textwidth}{!}{
\begin{tabular}{l|c|cccccccccccccccccc}
\midrule[1pt]    
 \multirow{2}{*}{\normalsize{Method}} &  \multirow{2}{*}{\normalsize{Backbone}}
 & \multicolumn{2}{c}{Car} & \multicolumn{2}{c}{Person} & \multicolumn{2}{c}{Bike} & \multicolumn{2}{c}{Curve} & \multicolumn{2}{c}{Car Stop} & \multicolumn{2}{c}{Guardrail} & \multicolumn{2}{c}{Color Cone} & \multicolumn{2}{c}{Bump}  
 & \multirow{2}{*}{\normalsize{mAcc}} & \multirow{2}{*}{\normalsize{mIoU}} \\
 
 \cline{3-18} 
     &  & Acc & IoU & Acc & IoU & Acc & IoU & Acc & IoU & Acc & IoU & Acc & IoU & Acc & IoU & Acc & IoU \\

\midrule[1pt] 
MFNet$_{\mathsf{17}}$~\cite{ha2017mfnet}  & -- & 77.2 & 65.9 & 67.0 & 58.9 & 53.9 & 42.9 & 36.2 & 29.9 & 19.1 & 9.9 & 0.1 & 8.5 & 30.3 & 25.2 & 30.0 & 27.7 & 45.1 & 39.7 \\	
ABMDRNet$_{\mathsf{21}}$~\cite{zhang2021abmdrnet}    & ResNet-50 & 94.3 & 84.8 & 90.0 & 69.6 & 75.7 & 60.3 & 64.0 &  45.1 & 44.1 & 33.1 & 31.0 & 5.1 &  61.7 & 47.4 & 66.2 & 50.0 & 69.5 &  54.8 \\
GMNet$_{\mathsf{21}}$~\cite{zhou2021gmnet}          & ResNet-50 & 94.1 & 86.5 & 83.0 & 73.1 & 76.9 & 61.7 & 59.7 & 44.0 & 55.0 & \underline{42.3} & \underline{71.2} & \underline{14.5} & 54.7 & 48.7 & 73.1 & 47.4 & 74.1 & 57.3 \\
FEANet$_{\mathsf{21}}$~\cite{deng2021feanet}         & ResNet-152 & 93.3 &87.8& 82.7&71.1& 76.7& 61.1 &65.5& 46.5 &26.6& 22.1& 70.8& 6.6& 66.6& 55.3& 77.3 &48.9& 73.2& 55.3\\
EGFNet$_{\mathsf{22}}$~\cite{zhou2022edge}          & ResNet-152 & \underline{95.8}&  87.6 &  89.0 & 69.8 &  80.6 & 58.8 & \underline{71.5} & 42.8 & 48.7 &  33.8 & 33.6 & 7.0 & 65.3 & 48.3 & 71.1 & 47.1 &  72.7 &  54.8 \\
MTANet$_{\mathsf{22}}$~\cite{zhou2022mtanet} 	& ResNet-152 & \underline{95.8} & 88.1 & \underline{90.9} & 71.5 & 80.3 & 60.7 & \textbf{75.3} & 40.9 & \underline{62.8} & 38.9 & 38.7 & 13.7 & 63.8 & 45.9 & 70.8 & 47.2 & 75.2 & 56.1 \\		
DSGBINet$_{\mathsf{22}}$~\cite{xu2022dual}  	& ResNet-152 & 95.2 & 87.4 & 89.2 & 69.5 & \textbf{85.2} & 64.7 & 66.0 & 46.3 & 56.7 & \textbf{43.4} & 7.8 & 3.3 & \underline{82.0} & \textbf{61.7} & 72.8 & 48.9 & 72.6 & 58.1 \\
FDCNet$_{\mathsf{22}}$~\cite{zhao2022feature}  	& ResNet-50 & 94.1 & 87.5 & \textbf{91.4} & 72.4 & 78.1 & 61.7 & 70.1 & 43.8  & 34.4 & 27.2 & 61.5 & 7.3 & 64.0 & 52.0 & 74.5 & 56.6 & 74.1 & 56.3 \\	
MFTNet$_{\mathsf{22}}$~\cite{zhou2022multispectral} & ResNet-152	& 95.1 &	87.9 	&85.2 &	66.8 &	83.9 &	64.6 &	64.3 &	47.1 &	50.8 &	36.1 &	45.9 &	8.4 &62.8 &	\underline{55.5} &	73.8 &	\underline{62.2} &74.7 &57.3 \\
LASNet$_{\mathsf{23}}$~\cite{li2022rgb}		 & ResNet-152 &   94.9 & 84.2 & 81.7 & 67.1 & 82.1 & 56.9 &  70.7 & 41.1 &56.8 & 39.6 &  59.5 & \textbf{18.9} & 58.1&  48.8 & 77.2 & 40.1 & \underline{75.4} & 54.9    \\

EAEFNet$_{\mathsf{23}}$~\cite{liang2023explicit} & ResNet-152 &	95.4 &	87.6 &	85.2 &	72.6 	& 79.9 	&63.8 	&70.6 &	48.6 &	47.9 &	35.0 &	62.8 	&14.2 &	62.7 &	52.4 &	71.9 &	58.3 &	75.1 &	58.9 \\
CAINet$_{\mathsf{24}}$~\cite{ying2024semantic} & MobileNet-V2 & 93.0	& 88.5	& 74.6	& 66.3	& \textbf{85.2}	& \textbf{68.7}	& 65.9	& \textbf{55.4}	& 34.7	& 31.5	& 65.6	& 9.0	& 55.6	& 48.9	& \textbf{85.0}	& 60.7	& 73.2	& 58.6 \\ 
CMX-B2$_{\mathsf{23}}$~\cite{zhang2023cmx} 	& MiT-B2$^\dagger$ & - & 89.4 & - & 74.8 & - & 64.7 & - & 47.3 & - & 30.1 & - & 8.1 & - & 52.4 & - & 59.4 & - & 58.2 \\
CMX-B4$_{\mathsf{23}}$~\cite{zhang2023cmx} 	& MiT-B4$^\dagger$ & -&\underline{90.1}& -& \underline{75.2}& -& 64.5& -& 50.2& - &35.3& -& 8.5& - &54.2& -& 60.6& - &59.7\\
CMNeXt$_{\mathsf{23}}$~\cite{zhang2023delivering} 	& MiT-B4$^\dagger$ & -&\textbf{91.5}& -& \textbf{75.3}& -& \underline{67.6}& -& 50.5& - &40.1& -& 9.3& - &53.4& -& 52.8& - &\underline{59.9}\\
SegMiF$_{\mathsf{23}}$~\cite{liu2023multi} &MiT-B3$^\dagger$&\textbf{96.3}& 87.8& 89.6& 71.4& 81.2& 63.2& 63.5& 47.5& \textbf{66.7}& 31.1& -& -& \textbf{85.3}& 48.9& \underline{84.8}& 50.3& 74.8& 56.1\\

\rowcolor{gray!20}
\textbf{SGFNet (Ours)}& ResNet-152 &	93.6 &	88.0 &	83.7 &	74.5 &	80.8 &	63.3 &	63.6 &	\underline{50.6} &	52.1 	&37.7 	&\textbf{74.9} &	10.8 &	63.0 &	53.7 &	77.9 &	\textbf{62.5} &	\textbf{76.2} &	\textbf{60.1} \\

\toprule[1pt]
\end{tabular}
}
\vspace{-15pt}
\end{table*}
\begin{table*}[t!]
  \centering
  \small
  \renewcommand{\arraystretch}{1.1}
  \renewcommand{\tabcolsep}{3mm}
  \caption{
  \textbf{Quantitative performance comparisons with the state-of-the-art methods on the PST900~\cite{shivakumar2020pst900} dataset}. The best results for each metric are in \textbf{bold}, while the second-best results are \underline{underlined}.}
  \label{tab:pst}
\resizebox{\linewidth}{!}{
\begin{tabular}{l|c|cccccccccccc}
\midrule[1pt]    
 \multirow{2}{*}{\normalsize{Method}} &\multirow{2}{*}{\normalsize{Backbone}} 
 & \multicolumn{2}{c}{Background} & \multicolumn{2}{c}{Hand-Drill} & \multicolumn{2}{c}{Backpack} & \multicolumn{2}{c}{Extinguisher} & \multicolumn{2}{c}{Survivor} 
 & \multirow{2}{*}{\normalsize{mAcc}} & \multirow{2}{*}{\normalsize{mIoU}} \\
 
 \cline{3-12} 
    & & Acc & IoU & Acc & IoU  & Acc & IoU & Acc & IoU & Acc & IoU \\
	     
\midrule[1pt] 
RTFNet$_{\mathsf{19}}$~\cite{sun2019rtfnet}  & ResNet-50 & 99.78 & 99.02 & 7.79 & 7.07 & 79.96 & 74.17 & 62.39 & 51.93 & 78.51  &  70.11 & 65.69 & 60.46 \\
GMNet$_{\mathsf{21}}$~\cite{zhou2021gmnet}   & ResNet-50 & \underline{99.81} & 99.44 & 90.29 & \textbf{85.17} & 89.01  & 83.82 & 88.28 & 73.79 & 80.65 & \underline{78.36} & 89.61 & 84.12 \\
MTANet$_{\mathsf{22}}$~\cite{zhou2022mtanet}   & ResNet-152    & - & 99.33 & - & 62.05 & - & \textbf{87.50} & - & 64.95 & - & \textbf{79.14} & - & 78.60 \\
DSGBINet$_{\mathsf{22}}$~\cite{xu2022dual}  & ResNet-152	& 99.73 & 99.39 & \textbf{94.53} & 74.99 & 88.65 & 85.11 & \underline{94.78} & 79.31 & \underline{81.37} & 75.56 & \underline{91.81} & 82.87 \\
    LASNet$_{\mathsf{23}}$~\cite{li2022rgb}	& ResNet-152	 & 99.77 & \underline{99.46} & 92.36 & 77.75 & \underline{90.80}  & 86.48 & 91.81 & \underline{82.80} & \textbf{83.43} & 75.49 & 91.63 & \underline{84.40} \\
\rowcolor{gray!20}
\textbf{SGFNet (Ours)} & ResNet-152 & \textbf{99.86}  & \textbf{99.52} &	\underline{92.62} &	\underline{79.89} &	\textbf{90.84} &	\underline{86.90} &	\textbf{94.99} &	\textbf{83.27} &	81.20 &	77.24 &	\textbf{91.90} &	\textbf{85.37} \\

\toprule[1pt]
\end{tabular}
}
\vspace{-10pt}
\end{table*}

\section{Experiments}
\vspace{-5pt}
\subsection{Experimental Settings}
\textbf{Datasets and Metrics}. Following the literature~\cite{liang2023explicit,zhou2021gmnet}, we use the MFNet~\cite{ha2017mfnet} and PST900~\cite{shivakumar2020pst900} datasets for benchmarking different RGB-T segmentation methods. The MFNet dataset consists of 1,569 pairs of RGB and thermal images, each with a resolution of $480\times 640$. 
We follow the splitting scheme~\cite{liang2023explicit,zhou2021gmnet} to use 50\% of the data for training, 25\% for validation, and 25\% for testing. 
We also evaluate our SGFNet on the PST900~\cite{shivakumar2020pst900} dataset, which consists of 894 matched RGB-T image pairs with pixel-level annotations for 4 classes. 
The resolution of the images in this dataset is $720\times 1280$. Of these, 597 image pairs are used for training, and the remaining 297 pairs are used for testing. 
Following previous work~\cite{sun2019rtfnet,ha2017mfnet}, we evaluate the quantitative performance of each method by mean accuracy (mAcc) and mean intersection over union (mIoU) metrics.

\textbf{Implementation Details}. We use the ImageNet pre-trained ResNet-152~\cite{he2016deep} model as the visual encoder for both modalities. We also adhere to the data augmentation protocol used in previous work~\cite{liang2023explicit}, incorporating random flip and random crop in our experiments. 
In both Equations~(\ref{eq:loss1}) and (\ref{eq:loss2}), the loss function is defined as the sum of the Dice~\cite{milletari2016v} loss and the SoftCrossEntropy loss. The size of the cosine bases is empirically set to $(H', W') = (7, 7)$.
In our experiments, we train our model for 100 epochs, initializing the learning rate at 0.02. The batch size is set to 2. An exponential scheduler is utilized to decrease the learning rate by a factor of 0.95 of its value per epoch. All the experiments are conducted on a single NVIDIA RTX 6000 Ada GPU with 48GB memory.

\begin{figure*}[t]
\centering
\includegraphics[width=\linewidth]{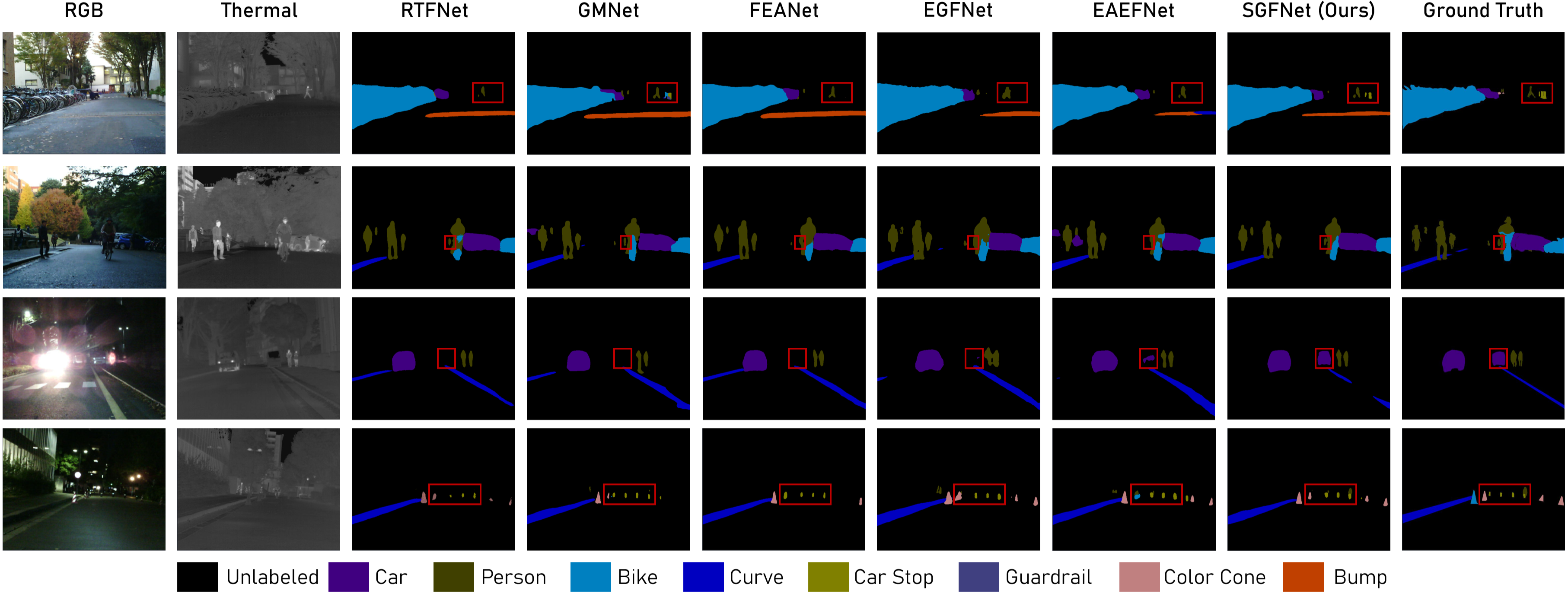}
\vspace{-19pt}
\caption{\revise{\textbf{Qualitative comparisons on the MFNet~\cite{ha2017mfnet} dataset}. We show the segmentation results for four test instances as examples: two captured during daytime (top) and the other two at nighttime (bottom). For easier comparison, the red boxes highlight specific areas of interest. Our proposed SGFNet demonstrates superior segmentation accuracy under various illumination conditions.}}
\label{fig:qualitative}
\vspace{-17pt}
\end{figure*}

\vspace{-5pt}
\subsection{Results and Discussions}
\noindent \textbf{Quantitative Results}. Table \ref{tab:mfnet} shows the performance comparisons of our SGFNet against other state-of-the-art methods on MFNet~\cite{ha2017mfnet}.
Our proposed SGFNet demonstrates superior segmentation accuracy, notably surpassing existing methods in mAcc and mIoU metrics. Specifically, compared to EAEFNet~\cite{liang2023explicit}, our model exhibits a performance gain of 1.1\% in mAcc and 1.2\% in mIoU.
We also compare our SGFNet with CMX~\cite{zhang2023cmx} and CMNeXt~\cite{zhang2023delivering}, which utilize the more advanced SegFormer~\cite{xie2021segformer} backbone. Remarkably, our SGFNet still outperforms these methods in the mIoU metric while using only a ResNet-152 backbone, highlighting the effectiveness of our designed SGF module.
Our SGFNet also delivers more accurate segmentation in challenging classes, \textit{e.g.}, achieving leading performance in $\mathsf{Curve}$ and $\mathsf{Bump}$ classes in terms of IoU, and the best performance in the $\mathsf{Guardrail}$ class in terms of accuracy. 

Moreover, we also report the performance of different methods on PST900~\cite{shivakumar2020pst900} in Table \ref{tab:pst}. Our proposed SGFNet achieves competitive performance, surpassing the second-best LASNet~\cite{li2022rgb} by 0.97\% in mIoU, and significantly outperforms other methods by even greater margins. 
These results further validate the effectiveness of our proposed SGF module in RGB-T segmentation tasks.

\textbf{Qualitative Results}. To provide a comprehensive insight into the effectiveness of SGFNet, we present a visual comparison of segmentation maps generated by our SGFNet and five representative state-of-the-art methods: RTFNet~\cite{sun2019rtfnet}, GMNet~\cite{zhou2021gmnet}, FEANet~\cite{deng2021feanet}, EGFNet~\cite{zhou2022edge}, and EAEFNet~\cite{liang2023explicit}. 
In the daytime scenes, our SGFNet accurately locates and segments small target objects. For instance, in the first daytime scene, where other methods struggle to detect the car stop, our SGFNet provides an accurate prediction. Additionally, in the second scene, our SGFNet successfully identifies the person in the background.
In the more challenging nighttime scenarios, where most objects are barely visible in the RGB images, our SGFNet can still successfully detect the car in over-exposed conditions and identify the person occluded by the car stop. 



\vspace{-5pt}
\subsection{Ablation Studies and Further Analysis}
\begin{table}[t]
\renewcommand{\arraystretch}{0.9}
\setlength\tabcolsep{10pt}
\centering
\caption{\textbf{Ablation studies of different algorithm components on MFNet~\cite{ha2017mfnet}}. In the table, SFE, SCA, GSA, and DS refer to spectral-aware feature enhancement, spectral-aware channel attention, global cross-modal spatial attention, and deep supervision, respectively.}
\label{tab:ablation}
\resizebox{\linewidth}{!}{
\begin{tabular}{c|cccc|cc}
\toprule
\#&SFE & SCA & GSA  & DS & mAcc ($\Delta$) & mIoU ($\Delta$)\\ \midrule
1&\XSolidBrush & \XSolidBrush &  \XSolidBrush   & \XSolidBrush &  73.1 (\textcolor{MidnightBlue}{0.0}) &  54.9 (\textcolor{MidnightBlue}{0.0}) 
 \\
\revise{2}&\revise{\XSolidBrush} & \revise{\XSolidBrush} &  \revise{\XSolidBrush}   &  \revise{\Checkmark} &  73.5 (\textcolor{MidnightBlue}{+0.4}) &  55.6 (\textcolor{MidnightBlue}{+0.7})
 \\
3&\XSolidBrush & \XSolidBrush &  \Checkmark   &  \XSolidBrush &  73.7 (\textcolor{MidnightBlue}{+0.6}) &  55.8 (\textcolor{MidnightBlue}{+0.9})
 \\
4&\XSolidBrush& \XSolidBrush& \Checkmark &   \Checkmark &  73.9 (\textcolor{MidnightBlue}{+0.8}) &  56.7 (\textcolor{MidnightBlue}{+1.8})
 \\
5&\Checkmark &   \Checkmark& \XSolidBrush& \XSolidBrush & 74.3 (\textcolor{MidnightBlue}{+1.1}) & 58.2 (\textcolor{MidnightBlue}{+3.3})
   \\
6&\Checkmark& \Checkmark& \XSolidBrush & \Checkmark &  74.8 (\textcolor{MidnightBlue}{+1.7}) &  58.5 (\textcolor{MidnightBlue}{+3.6})
\\ 
7&\Checkmark& \Checkmark& \Checkmark &  \XSolidBrush &  75.8 (\textcolor{MidnightBlue}{+2.7}) & 59.6 (\textcolor{MidnightBlue}{+4.7})
\\ 
\rowcolor{gray!20}
8&\Checkmark &\Checkmark  & \Checkmark & \Checkmark & \textbf{76.2} (\textcolor{MidnightBlue}{+3.1}) &  \textbf{60.1} (\textcolor{MidnightBlue}{+5.2})
\\ 
\bottomrule
\end{tabular}
}
\vspace{-13pt}
\end{table}
\noindent 
\textbf{Effectiveness of Different Components}. To systematically analyze the impact of different components in our SGFNet, we conduct an ablation study on the MFNet~\cite{ha2017mfnet} dataset. The results in Table \ref{tab:ablation} illustrate the effects of various combinations of components. 
We have the following key observations: (1) The inclusion of the spectral-aware feature enhancement and channel attention provide a notable 3.3\% improvement in mIoU; (2) Incorporating global cross-modal spatial attention and deep supervision into the model further enhances its performance by 1.4\% and 0.5\% mIoU, respectively; (3) Our full SGFNet, with all four algorithm components, achieves the highest accuracy of 76.2\% mAcc and 60.1\% mIoU.

\begin{table}[t]
\renewcommand{\arraystretch}{1.0}
\centering
\caption{\revise{\textbf{Efficiency comparison on the MFNet~\cite{ha2017mfnet} dataset}. We present the GFLOPs and \#Params of different models.}}
\label{tab:efficiency}
\resizebox{\linewidth}{!}{
\begin{tabular}{l|cccc}
\toprule
Method  & FLOPs/G $\downarrow$ & \#Params/M $\downarrow$ & mAcc $\uparrow$ & mIoU $\uparrow$ \\
\midrule
RTFNet$_{\mathsf{19}}$~\cite{sun2019rtfnet}   & 337.46 & 254.51 & 63.1 &53.2\\
ABMDRNet$_{\mathsf{21}}$~\cite{zhang2021abmdrnet}      & 250.51 & \textbf{139.24} & 69.5 & 54.8\\
FEANet$_{\mathsf{21}}$~\cite{deng2021feanet}     & 337.67 & 255.21 & 73.2 & 55.3\\
EAEFNet$_{\mathsf{23}}$~\cite{liang2023explicit}     & \textbf{200.97} & \underline{147.20} & 75.1 & 58.9\\
\rowcolor{gray!20}
\textbf{SGFNet (Ours)} & \underline{249.25} & 163.99 & \textbf{76.2} & \textbf{60.1}\\ 
\bottomrule
\end{tabular}
}
\vspace{-15pt}
\end{table}

\revise{
\textbf{Computational Complexity}. Table~\ref{tab:efficiency} presents a comparison of efficiency between our approach and existing methods, measured in terms of FLOPs and parameter count. The results demonstrate that our SGFNet not only achieves superior performance but also offers competitive computational efficiency. 
}

\begin{table}[t]
\renewcommand{\arraystretch}{1.0}
\setlength\tabcolsep{9pt}
\centering
\caption{\revise{\textbf{Impacts of different backbones}. We present the computational complexity and the corresponding MFNet~\cite{ha2017mfnet} accuracy achieved by our SGFNet with different ResNet backbones.}}
\label{tab:backbone}
\resizebox{\linewidth}{!}{
\begin{tabular}{l|cccc}
\toprule
Backbone  & FLOPs/G $\downarrow$ & \#Params/M $\downarrow$ & mAcc $\uparrow$ & mIoU $\uparrow$ \\
\midrule
ResNet-18   & \textbf{94.90} & \textbf{30.59} & 72.3 &55.6\\
ResNet-34   & 117.61 & 50.80 & 73.9 & 56.3\\
ResNet-50   & 156.50 & 93.32 & 73.6 & 57.4\\
ResNet-101  & 202.20 & 131.30 & 75.3 & 58.9\\
\rowcolor{gray!20}
ResNet-152 & 249.25 & 163.99 & \textbf{76.2} & \textbf{60.1}\\ 
\bottomrule
\end{tabular}
}
\vspace{-14pt}
\end{table}

\revise{
\textbf{Scaling Backbones}. In Table~\ref{tab:backbone}, we provide an analysis of the computational complexity along with the corresponding MFNet~\cite{ha2017mfnet} accuracy using various ResNet backbones. We observe a trade-off between efficiency and performance: larger backbones yield higher performance but also increase computational complexity.
}

\vspace{-3pt}
\section{Conclusion}
\looseness=-1 
In this work, we propose Spectral-aware Global Fusion Network (SGFNet) for RGB-T semantic segmentation. Specifically, we enhance the precision of segmentation by refining the fusion process of multi-modal features, focusing on amplifying the interactions among high-frequency features that contain distinct and modality-specific details. We also design spectral-aware feature enhancement and spectral-aware channel attention to facilitate cross-modal feature interaction in the spectral domain.
Furthermore, we develop a global cross-modal spatial attention module that computes the cross-attention across all pixel pairs, incorporating global spatial relationships for effective cross-modal feature fusion. Empirical evaluations show that SGFNet surpasses state-of-the-art methods on two standard RGB-T semantic segmentation datasets, MFNet and PST900.

\section{Acknowledgments} 
\sloppy
This work has been funded in part by Army Research Laboratory (ARL) award W911NF-23-2-0007 and W911QX-24-F-0049, DARPA award FA8750-23-2-1015, and ONR award N00014-23-1-2840.
\fussy
\bibliographystyle{IEEEbib}
\bibliography{reference}

\end{document}